# Identification of splicing edges in tampered image based on Dichromatic Reflection Model

SHEN ZHE[1], SUN PENG[2*], LANG YUBO[2], LIU LEI[3], PENG SILONG[4]

[1]Civil aviation college, Shenyang Aerospace University, Shenyang 110136, China
[2] Department of Criminal Science and Technique, Criminal Investigation Police University of China, Shenyang, 110035, China
[3]College of Information Science and Engineering, Northeastern University, Shenyang 110819, China
[4]Institute of Automation Chinese Academy of Sciences, Beijing, 100190,China
*Corresponding author: 6094079@qq.com

**Imaging is a sophisticated process combining a plenty of photovoltaic conversions, which lead to some spectral signatures beyond visual perception in the final images. Any manipulation against an original image will destroy these signatures and inevitably leave some traces in the final forgery. Therefore we present a novel optic-physical method to discriminate splicing edges from natural edges in a tampered image. First, we transform the forensic image from $RGB$ into color space of $S$ and $o_1 o_2$. Then on the assumption of Dichromatic Reflection Model，edges in the image are discovered by composite gradient and classified into different types based on their different photometric properties. Finally, splicing edges are reserved against natural ones by a simple logical algorithm. Experiment results show the efficacy of the proposed method.**

The method using image processing software to piece together parts of content from two or more images into a new composite image, is widely employed to present fraudulent evidence. Thus, it causes potential hazards in political, economic and forensic activities. At present, the most adopted methods to confirm a composite image can be summarized as follows: optics-related methods, device-related methods and encoder-related methods. However, all these methods have limitations. On one hand，primarily dependent on the periodical variation of DCT and the features of double JPEG image compressing, the encoder-related methods proposed in [1] is in most cases unlikely to effectively differentiate normal editing from vicious manipulations; On the other hand, current optics and device related methods are mainly based on the imaging variations such as light direction[2], light color[3]and pattern noise[4] in different parts of composite image so that these methods rest largely on manual handling or sample training to find out manipulation areas. Consequently, with no samples available, they could not be practically applied to single image inspection. The edge, an important sort of image characters which can reflect scenery light conditions, material properties, geometrical shapes and relative position of objects in a scene, plays an unusual role in scenery analysis and object detection in an image. Apart from natural edges, there are also splicing edges which are inevitably left by manipulations. If splicing edges in images can be correctly marked, then manipulations will be confirmed and their area will be located at the same time. The method to locate manipulations by detecting edges has been proposed previously in [5]. It analyzes gradient variation between splicing edges and natural edges, calculates derivative histogram entropy of every pixel in detected edges and then compares this derivative histogram entropy with a manual threshold to identify which are splicing edges. Although this method introduces a brand new perspective in composite image detecting, the low precision of locating manipulation area is at a disadvantage because of the high complexity of calculating each edge point and the susceptibility to noise. It has been proven that reflection color model is effective for separating dichromatic reflection components using a single image [6,7]. Though taking advantage of different sensitivity to natural image edges and splicing edges in different color space edge based on Dichromatic Reflection Model [8,9], our method carries out logical operations on binary images after the classification of image edges in $S$ and $o_1 o_2$ color space, and seek out splicing edges for a better application to locate manipulation area in composite image with only one single forensic image.

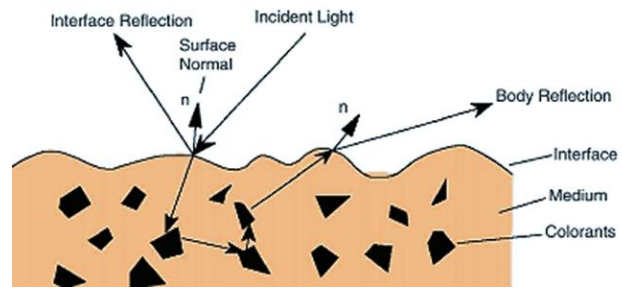

Fig.1. schematic diagram of Dichromatic Reflection Model

Shafer's Dichromatic Reflection Model in Fig.1 was the first physics-based model of reflection to separate different types of reflection on a single surface [10]. It considers an image of an infinitesimal surface patch of an inhomogeneous dielectric object and denotes the red, green and blue sensors with spectral sensitivities as $f_R(\lambda)$, and $f_G(\lambda)$, $f_B(\lambda)$ respectively. For an image of the surface patch illuminated by a SPD of the incident light denoted by $e(\lambda)$, the measured sensor values $C \in \{R,G,B\}$ are given by:

$$C = m_b(\mathbf{n},\mathbf{s})\int_\lambda f_C(\lambda)e(\lambda)c_b(\lambda)d\lambda + m_s(\mathbf{n},\mathbf{s},\mathbf{v})\int_\lambda f_C(\lambda)e(\lambda)c_s(\lambda)d\lambda \quad (1)$$

Where $f_C(\lambda) \in \{f_R(\lambda), f_G(\lambda), f_B(\lambda)\}$, $m_b$ and $m_s$ denote the geometric dependencies on the body and interface reflection component respectively. $n$, $s$ and $v$ denote the surface patch normal, direction of the illumination source, and direction of the viewer respectively. $\lambda$ is the wavelength, $c_b(\lambda)$ and $c_s(\lambda)$ are the Fresnel reflectance and surface albedo respectively. $e(\lambda)$ and $c_s(\lambda)$ respectively have a constant value (i.e. $e$ and $c_s$) on the assumptions of white or spectrally smooth illumination (i.e., approximately equal or smooth energy density for all wavelengths within the visible spectrum, can be simply named as quasi-white light) and the neutral interface reflection (NIR) model (i.e. has a constant value independent of the wavelength) [11]. For $C_w \in \{R_w, G_w, B_w\}$ giving the red, green and blue sensor response under the assumption of quasi-white light source, it can be further concluded that $k_C = \int_\lambda f_C(\lambda)c_b(\lambda)d\lambda$, and $\int_\lambda f_R(\lambda)d\lambda = \int_\lambda f_G(\lambda)d\lambda = \int_\lambda f_B(\lambda)d\lambda = f$, then the sensor values can be put forward by:

$$C_w = em_b(\mathbf{n},\mathbf{s})k_C + em_s(\mathbf{n},\mathbf{s},\mathbf{v})c_s f \quad (2)$$

Weijer et al. used the above reflection model to develop an algorithm for separating edges in an image as material edges (e.g. edges between objects and object-background), shadow/shading edges (e.g. edges caused by the shape or position of an object with respect to the light source) and specular edges (i.e. highlight) [12]. Corresponding to geometrical and photometric conditions within a scene, Maxwell et al. further distinguish edges as one of the following five classes: object edges, reflectance edges, shadow edges, specular edges, and occlusion edges [13]. The above methods all depend on the usage of color spaces, and Fig.2. shows a tampered image in different color spaces. In tampered images, in addition to natural edge types, there is another kind of edge named splicing edge which is essentially different from natural edges. The proposed method uses two photometric color spaces named $S$ and $o_1 o_2$, which are derived from the dichromatic reflection model and are proven to be insensitive to certain types of edges. $S$ is also known as saturation and can be calculated by:

$$S = 1 - \frac{\min(R,G,B)}{R+G+B} \quad (3)$$

Because $S$ corresponds to the radial distance from the color to the main diagonal in the RGB-color space, $S$ is an invariant for matte, dull surfaces illuminated by quasi-white light which defined

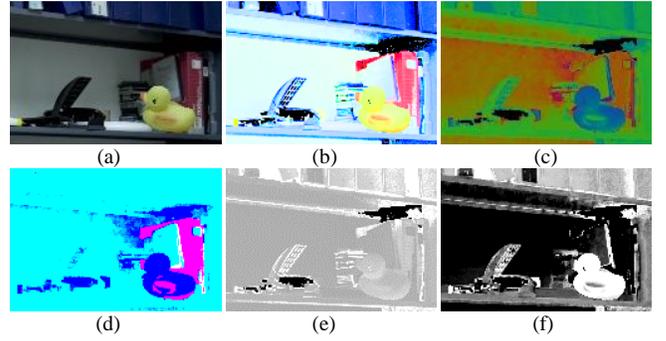

Fig.2. Splicing images in different color spaces.
(a) RGB  (b) C1C2C3  (c) L1L2L3  (d) O1O2  (e) S  (f) H.

before, then we go further by substituting $C_w = em_b(\mathbf{n},\mathbf{s})k_C$ in to (3):

$$S = 1 - \frac{\min(em_b(\mathbf{n},\mathbf{s})k_R, em_b(\mathbf{n},\mathbf{s})k_G, em_b(\mathbf{n},\mathbf{s})k_B)}{em_b(\mathbf{n},\mathbf{s})(k_R + k_G + k_B)}$$
$$= 1 - \frac{\min(k_R, k_G, k_B)}{k_R + k_G + k_B} \quad (4)$$

Where $k_C \in \{k_R, k_G, k_B\}$, (4) shows that $S$ only depend on the sensors and the surface albedo. Therefore edges corresponding to highlight and material are approximately easy to be detected in this color space. Further, we consider a two-dimensional opponent color space $o_1 o_2$ (leaving out the intensity component here) which is defined by:

$$o_1 = \frac{(R-G)}{2} \quad (5)$$

$$o_2 = \frac{B}{2} - \frac{R+G}{4} \quad (6)$$

For shiny surfaces, the above equations are formulated further by substituting (2) into (5), then we get (7) and (8):

$$o_1 = \frac{R_w - G_w}{2}$$
$$= \frac{em_b(\mathbf{n},\mathbf{s})k_R - em_b(\mathbf{n},\mathbf{s})k_G}{2} \quad (7)$$
$$= em_b(\mathbf{n},\mathbf{s})\frac{k_R - k_G}{2}$$

$$o_2 = \frac{B_w}{2} - \frac{R_w + G_w}{4}$$
$$= \frac{2em_b(\mathbf{n},\mathbf{s})k_B - em_b(\mathbf{n},\mathbf{s})k_R - em_b(\mathbf{n},\mathbf{s})k_G}{4} \quad (8)$$
$$= em_b(\mathbf{n},\mathbf{s})\frac{2k_B - k_R - k_G}{4}$$

Where $R_w = em_b(\mathbf{n},\mathbf{s})k_R + em_s(\mathbf{n},\mathbf{s},\mathbf{v})c_s f$, $G_w = em_b(\mathbf{n},\mathbf{s})k_G + em_s(\mathbf{n},\mathbf{s},\mathbf{v})c_s f$ and $B_w = em_b(\mathbf{n},\mathbf{s})k_B + em_s(\mathbf{n},\mathbf{s},\mathbf{v})c_s f$. From (7) and (8), it can be concluded that color space of $o_1 o_2$ is insensitive to highlight,. Furthermore, since the materials used for image splicing come from different imaging processes, they have obvious different photometric properties. The specific manifestation is that the surface normal $n$ and illumination direction $s$ of the medium surface usually change greatly at the splicing edge area.

According to the above formulas (7) and (8), we can find that the color space of $o_1o_2$ is sensitive to the changes of $n$ and $s$, so it should be sensitive to splicing edges in a composite image. Tab. 1. shows sensitivities of $n$ and $o_1o_2$ to various types of edges

Tab. 1. Overview of sensitivities of $n$ and $o_1o_2$ to various types of edges, where + means sensitive to edges, - means not sensitive to edges.

|       | Shading edge | Shadow edge | Highlight edge | Material edge | Splicing edge |
|-------|:---:|:---:|:---:|:---:|:---:|
| $O_1O_2$ | + | + | - | + | + |
| S | - | - | + | + | - |

Immediately after converting the image from $RGB$ to $S$ and $o_1o_2$, laplace operator is used to calculate the gradient of different channels of the image in color spaces of $S$ and $o_1o_2$ especially.

$$f_i' = f_i \otimes h, \quad h = \begin{bmatrix} 0 & 1 & 0 \\ 1 & -4 & 1 \\ 0 & 1 & 0 \end{bmatrix} \quad (9)$$

Where $f_i$ denotes $ith$ channel of $S$ and $o_1o_2$, i.e. $i \in \{S, o_1, o_2\}$, $h$ represents laplace operator, and $f_i'$ denotes the convolution of $f_i$ and $h$, representing the edges detected in the $ith$ channel of the image in $S$ and $o_1o_2$ color space. Since laplace operator can largely enhance the gradient difference between the object and the background, the edge pixel can be effectively highlighted. Considering that some color spaces have several dimensions, such as $o_1o_2$, here we introduce a composite gradient $\nabla F$ which can be calculated by:

$$\nabla F(x,y) = \sqrt{\sum_{i=1}^{N} f_i'(x,y)^2} \quad (10)$$

Where $N$ is the number of color space channels. $\nabla F$ can further improve the difference between a pixel and its neighbor pixels. A pixel with a comparative large $\nabla F$ indicates that there is a big difference between its adjacent pixels, so it is likely to be an edge. Some studies have suggested that the pixel gradient can be identified as edge pixel when it is more than 3 standard deviations[8]:

$$\nabla C(x,y) = \begin{cases} 1 & \text{if } \nabla F(x,y) > 3\sigma_{\nabla F} \\ 0 & \text{else} \end{cases} \quad (11)$$

Where $\sigma_{\nabla F}$ is the standard deviation of $\nabla F$. According to the value of $\nabla C(x,y)$, the detected pixels can either be classified into edge or not. After that, $\nabla F$ in (11) is transformed into edge maps as $\nabla C_S$ and $\nabla C_{o_1o_2}$, respectively corresponding to color space. As mentioned in Tab1, color space $o_1o_2$ is sensitive to splicing edges, and $S$ is on the contrary. Finally, let $\nabla C_{Sp}$ represent detected splicing edges, and a novel color edge taxonomy is presented. Based on this taxonomy the splicing pixels can be better identified by the logical combination of $\nabla C_S$ and $\nabla C_{o_1o_2}$. The logical algorithm is listed below:

$$\text{IF } \nabla C_{o_1o_2} = 1 \text{ AND } \nabla C_S = 0$$
$$\text{THEN } \nabla C_{Sp} = 1$$
$$\text{ELSE } \nabla C_{Sp} = 0$$

Similar to previous works, the performance of the proposed algorithm was first validated on popular benchmark dataset named Columbia Image Splicing Detection Evaluation Dataset[14], which consists of 180 original images and 183 tampered images. While the tampered images are spliced automatically, almost all splicing edges are not consistent with the scene. In order to get a better visual performance, we construct a new splicing dataset of semantic forgery with original images in Columbia dataset. In our new dataset, splicing edges are cut along with material edges to simulate a more realistic image mainly beyond visual inspection. As a result, the visual inspection experiments compared with [6] are shown in Fig.3, from which, we can find that our method can get a better visual performance than [6] across four tampered samples with diversified objects spliced in different scenes. Meanwhile detected edges in $S$ and $o_1o_2$ are also presented for clarity. It can be observed that the detected edges vary drastically among different color space, and this variation are also correlated with surface texture, material and illumination etc..

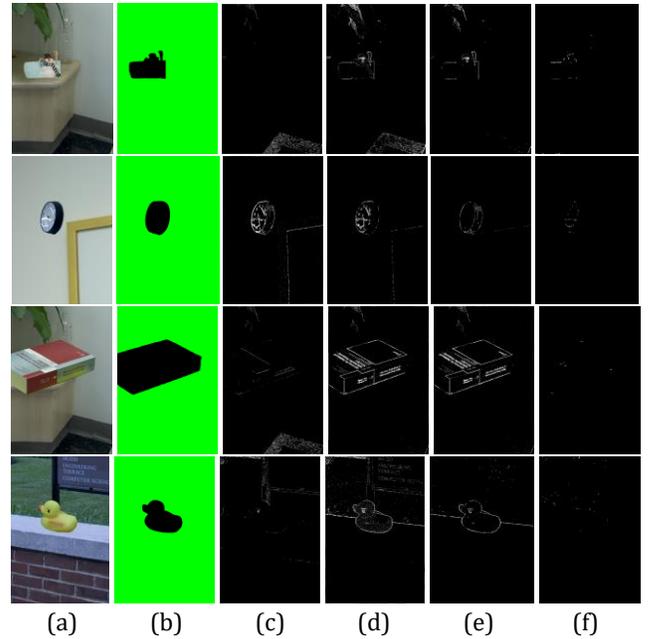

(a)　　(b)　　(c)　　(d)　　(e)　　(f)

Fig.3. Visual evaluation of detected results of our method comparing with [6] on a novel dataset comes from Columbia dataset .Images in turn are (a)splicing image (b)MASK (c)edges in S (d) edges in O1O2 (e)our method (f) [6]

Next, experiments are carried out for qualitative evaluations on the proposed method. Here we use a quantitative index named link precision denoted as $F1$ which have been universally accepted by other researchers [15,16]. It combines precision and recall in a unified framework to achieve an equalized evaluation. The quantitative evaluation results are shown in Tab.2. , in which a comprehensive comparison is also provided between our method and [6].

Tab. 2. Quantitative evaluation of the proposed method compares with [6] on Columbia dataset.

|  | $F1_{Max}$ | $F1_{Mean}$ | $F1_{Median}$ | $BR_{Mean}$ |
|---|---|---|---|---|
| The proposed method | 0.8823 | 0.4947 | 0.5125 | 0.6742 |
| [6] | 0.8059 | 0.4232 | 0.4213 | 0.3400 |

In the above Tab.1. $F1_{Max}$, $F1_{Mean}$ and $F1_{Median}$ are statistics of $F1$. And as a supplement, $BR_{Mean}$ is used to denote average of the boundary recall of splicing edges [17]. It is means that only if BR is large enough, the detected pixel is a meaningful splicing edge pixel.

$$B = \begin{cases} 1 & \text{if} \quad BR \geq \theta \\ 0 & \text{else} \end{cases} \quad (12)$$

Where $B$ is a logical variable used to determine whether the pixel is an edge pixel, and $\theta$ is a preset threshold. Let $Ed_{SD}$ denotes the number of pixels of detected splicing edge in a splicing image, $Ed_{OD}$ denotes the number of pixels of detected edge in an original image, as well as $Ed_S$ is the number of pixels of the full splicing edge, and $N_O$ is the number of pixels of the whole image. And considering both splicing samples and original ones, we another threshold $\alpha$ defined by them to evaluate the final detected performance [3]. Fig.4 is an exhibition of quantitative evaluation, where (a) shows the variation of $F1$ when $\theta$ goes from 0 to 1, and (b) shows ROC curves combined with True Positive Rate against False Positive Rate, where $\theta = 0.3$. The proposed method and [6] (FZ) are presented together for a quantitative comparison. From Fig.4 we can observe that the proposed method have a better performance than [6].

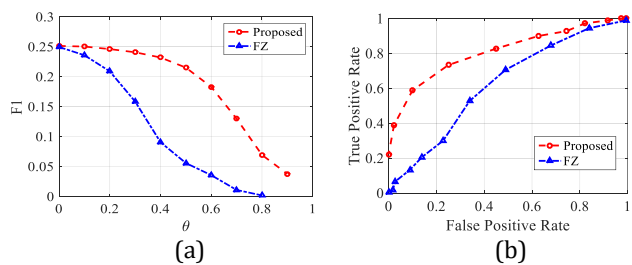

Fig.4. Quantitative evaluation of our method comparing with [6] on Columbia dataset. (a) F1 curve obtained by θ varying from 0 to 1 (b) ROC curve obtained by α varying from 0 to 1

In conclusion, we provide a novel method for splicing manipulation detection based on Dichromatic Reflection Model without any training samples. Due to its simplicity and excellent quantitative performance, the proposed technique should be suitable for a large number of forensic applications with only one single image presented.

**Funding.** Natural National Science Foundation of China (NSFC) (61307016). National Key R&D Program (2017YFC0822204). National Engineering Laboratory of Evidence Traceability Technology (2017NELKFKT09).
**Disclosures.** The authors declare no conflicts of interest.